\documentclass[11pt]{article}

\usepackage[preprint]{acl}

\usepackage{times}
\usepackage{latexsym}
\usepackage{float}
\usepackage{enumitem}
\usepackage{graphicx} % in preamble
\usepackage{booktabs}
\usepackage{tabularx}
\usepackage{siunitx}
\usepackage{hyperref}
\usepackage[T1]{fontenc}
\usepackage[utf8]{inputenc}

\usepackage[T1]{fontenc}
\usepackage[utf8]{inputenc}

\usepackage{microtype}

\usepackage{inconsolata}

\usepackage{graphicx}
\usepackage{booktabs}
\usepackage{hyperref}

\title{Lights, Camera, Consistency: A Multistage Pipeline for Character-Stable AI Video Stories}

\author{
  \textbf{Chayan Jain\textsuperscript{1}},
  \textbf{Rishant Sharma\textsuperscript{1}},
  \textbf{Archit Garg\textsuperscript{1}},
  \textbf{Ishan Bhanuka\textsuperscript{1}},
  \\
  \textbf{Pratik Narang\textsuperscript{1},}
  \textbf{Dhruv Kumar\textsuperscript{1}}
\\
  \textsuperscript{1}BITS Pilani, Pilani Campus, India\\
  \small{
  \textbf{Correspondence:} \href{mailto:dhruv.kumar@pilani.bits-pilani.ac.in}{chayanhjain@gmail.com}
 }
}

\begin{document}
\maketitle
% --- ABSTRACT SECTION ---
\begin{abstract}
Generating long, cohesive video stories with consistent characters is a significant challenge for current text-to-video AI. We introduce a method that approaches video generation in a filmmaker-like manner. Instead of creating a video in one step, our proposed pipeline first uses a large language model to generate a detailed production script. This script guides a text-to-image model in creating consistent visuals for each character, which then serve as anchors for a video generation model to synthesize each scene individually. Our baseline comparisons validate the necessity of this multi-stage decomposition; specifically, we observe that removing the visual anchoring mechanism results in a catastrophic drop in character consistency scores (from 7.99 to 0.55), confirming that visual priors are essential for identity preservation. Furthermore, we analyze cultural disparities in current models, revealing distinct biases in subject consistency and dynamic degree between Indian vs Western-themed generations.
\end{abstract}
\section{Introduction}

The domain of text-to-video generation represents a frontier at the intersection of natural language processing (NLP) and computer vision, where realistic and coherent video sequences are synthesized directly from textual descriptions. The domain has rapidly advanced with the advent of diffusion models \citep{ho2020denoising-nips}. These models excel at generating short, high-quality video clips from textual prompts, demonstrating exceptional capabilities in generating dynamic scenes and motions \citep{singer2022makeavideo, ho2022imagenvideo}.

Recent research in text-to-video (T2V) generation has primarily focused on improving visual quality and temporal coherence within short, continuous clips, leading to powerful models such as Google's Veo \cite{google-veo} and OpenAI's Sora \cite{openai-sora}. Despite these advances, the prevailing T2V paradigm remains largely single-shot, where a single prompt produces one uninterrupted clip. While effective for short sequences, this approach is inherently limited in its ability to generate structured narratives involving multiple scenes, recurring characters, and coherent plot progression. 

Existing literature largely reflects this clip-centric focus, emphasizing foundational model training \citep{ho2022imagenvideo}, efficient fine-tuning methods \citep{wu2023tuneavideo}, and benchmark development for evaluating per-clip fidelity and text–video alignment \citep{chen2023evalcrafter}. As a result, a significant gap remains in the development of holistic frameworks capable of producing long-form, multi-scene stories with consistent characters and narrative structure. To address this limitation, we propose a system that automates the entire filmmaking pipeline—from script generation to final video composition—enabling the creation of coherent, story-driven videos from a single prompt.

To bridge this gap, we introduce a novel inference-only pipeline for generating narrative-driven videos from a single text prompt. Unlike frameworks relying on probabilistic agent simulations, our approach employs structured deterministic control via JSON blueprints to enforce precise adherence to a script. This methodology mimics a traditional filmmaking workflow but is optimized for computational efficiency. First, we utilize a Large Language Model (LLM) to generate a script with strict scene constraints. Second, we introduce a zero-shot ``Asset-First'' mechanism that decouples character design from scene generation; this ensures visual consistency through direct conditioning, eliminating the need for expensive fine-tuning or LoRA training. Third, a scene synthesis module utilizes a ``Temporal Bridge'' to explicitly model the flow of time, using the final frame of a previous scene to visually condition the next. Finally, a merging module produces the final output with synchronized audio.

Our evaluation highlights three critical insights regarding this architecture. First, we demonstrate that the generated image-to-image (I2I) seed frame is the architectural linchpin for consistency; removing this visual anchor results in a catastrophic drop in character consistency scores (from 7.99 to 0.55), proving that a character model alone is insufficient for identity retention. Second, we identify a distinct performance gap between visual and narrative fidelity: while the model excels at Prompt Adherence (Score: 4.47), it struggles with strict Script Adherence (Score: 3.49), indicating that current architectures prioritize style transfer over complex narrative sequencing. Finally, our bias analysis uncovers a ``Subject-World Decoupling'' phenomenon. We find that while the model maintains non-Western (e.g., Indian) character identities effectively, the physical coherence of the surrounding environment degrades significantly under high-motion stress compared to Western contexts, highlighting a specific robustness failure in diverse physical simulations.

\section{Related Work}

\subsection{Multi-Agent Frameworks}
% \subsection{Structured Inference versus Agent-Based and Training-Heavy Paradigms}
Current multi-agent frameworks often decompose video generation into sub-tasks mimicking film production. However, our methodology distinguishes itself from these approaches by prioritizing structured, deterministic inference over probabilistic simulation or computationally intensive fine-tuning.

Unlike DreamFactory \cite{xie2024dreamfactory}, which relies on probabilistic social negotiation between roleplaying agents (e.g., Director vs. Screenwriter) to derive scene parameters, our approach utilizes deterministic JSON blueprints. This shifts the paradigm from simulating a conversation to engineering a control flow. Furthermore, while DreamFactory relies on text-based keyframe iteration, we decouple character design from scene generation. By employing an asset-first strategy—generating explicit character sheets before scene synthesis—we ensure visual consistency through direct visual conditioning rather than semantic descriptions alone.

In contrast to StoryAgent \cite{hu2024storyagent}, which necessitates computationally expensive fine-tuning (e.g., LoRA-BE) to embed character identity into model weights, our framework offers a scalable, inference-only solution. We achieve consistency through a zero-shot ``temporal bridge,'' where the final frame of a preceding scene visually conditions the initial frame of the subsequent shot. This explicitly models the linear flow of time without the overhead of retraining models for every new narrative.

Finally, compared to AniMaker \cite{shi2025animaker}, which employs computationally heavy Monte Carlo Tree Search (MCTS) for quality control, our pipeline emphasizes efficiency and low latency. By leveraging a linear, structured workflow grounded by LLM-generated anchor frames, we avoid the latency of iterative search algorithms while maintaining multimodal completeness, including the integration of audio and scoring absent in many visual-only pipelines like Mora \cite{yuan2024mora}.

% --- Placeholder for the pipeline figure ---
\begin{figure*}[!t]
  \centering
  \includegraphics[width=0.95\textwidth]{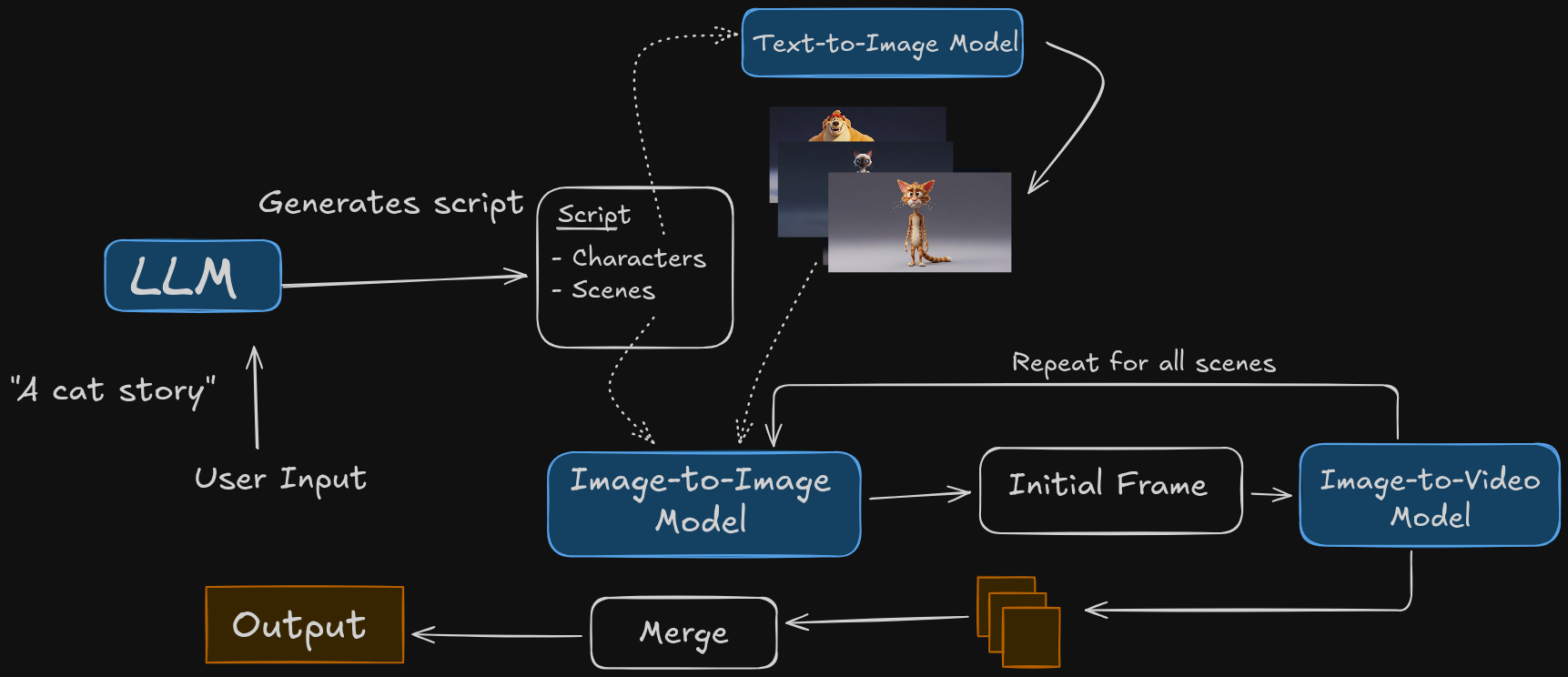} 
  \caption{The Proposed Pipeline. A high-level user prompt initiates an LLM to generate a detailed script and character descriptions. A Text-to-Image model creates visual assets for characters. For each scene, an Image-to-Image model generates an initial frame, which then guides an Image-to-Video model to synthesize the scene clip. All clips are then merged into the final output video.}
  \label{fig:pipeline}
\end{figure*}

\subsection{Video Generation}
Driven by the success of diffusion models \cite{ho2020denoising-nips}, video generation has evolved from simple text-to-video (T2V) tasks to complex image-to-video (I2V) synthesis. Pioneering zero-shot frameworks like Text2Video-Zero \cite{khachatryan2023text2videozero} demonstrated that pretrained text-to-image models could generate video without additional training. Subsequent works have introduced temporal self-attention and LLM-guided frame prompts to improve coherence \cite{wu2022tuneavideo, qi2023text}. Notably, DirecT2V \cite{hong2024direct2v} leverages LLMs to generate time-varying prompts for better controllability. However, these methods primarily operate on a per-clip basis, often struggling to maintain narrative and character consistency when extending beyond a single short sequence.

In the domain of Image-to-Video (I2V), architectures like Dynamic-I2V \cite{liu2025dynamici2v} represent the state-of-the-art, utilizing multimodal large language models (MLLMs) and conditional adapters to synthesize video from single images. While these models excel at animating static inputs with high motion fidelity, they are typically designed for single-shot animation rather than sequential storytelling. Our work repurposes these I2V capabilities, utilizing them not just for motion generation, but as a ``temporal bridge'' mechanism where the final frame of one scene serves as the visual anchor for the next.

Recent benchmarks such as VBench \cite{huang2024vbench} and DEVIL \cite{liao2024devil} have shifted evaluation focus toward balancing motion realism with frame-wise quality. Despite these advances, a gap remains in holistic frameworks capable of producing long-form narratives. Unlike existing approaches that rely on probabilistic generation or complex fine-tuning, our pipeline introduces a deterministic, blueprint-driven workflow to enforce character and narrative consistency across multiple generated clips.
\section{Methodology}

To address the challenge of generating long-form structured videos, we propose a multistage pipeline that breaks down the generation task into a sequence of sub-problems. Our approach mimics a traditional filmmaking workflow, proceeding from a concept to scriptwriting, character design, scene rendering, and final composition.

\subsection{Stage 1: Script and Blueprint Generation}

The pipeline \hyperref[fig:pipeline]{(Figure~\ref*{fig:pipeline})} begins with a user prompt that outlines the desired story (e.g., ''A cat story''). The prompt is fed into a scriptwriting module consisting of a LLM such as Gemini 2.5 suite \citep{comanici2025gemini2p5}. We carefully engineer a prompt that instructs the LLM to act as a screenwriter and director which outputs a structured JSON object. This blueprint contains three main components:
\begin{itemize}[leftmargin=*]
    \item \textbf{Story Analysis \& Art Style}: Details defining the core themes, and a detailed \texttt{art\_style} object specifying the visual aesthetic, color palette, and references (e.g., ``Gritty Epic CGI'' inspired by ``Game of Thrones'').
    
    \item \textbf{Character Sheets}: A list of all characters in the story. Each character is defined by their name, role, and personality, detailed \texttt{physical\_description}, \texttt{visual\_notes}, and a complete character arc (e.g., ``a ginger tabby cat wearing a tiny trench coat and fedora'').
    
    \item \textbf{Scene Breakdown}: A sequential list of scenes. Each scene is defined by its duration, setting, a description of the \texttt{action}, a list of characters present, visual notes for cinematography, sound effects, and the emotional tone. This structured output serves as the blueprint for the next stages, which ensures a clear narrative and well-described features.
\end{itemize}

\subsection{Stage 2: Character Visualisation}
For each character identified in the production blueprint, this module uses a Text-to-Image (T2I) Model to generate a set of consistent images. The prompt for the T2I model is meticulously constructed from the character's \texttt{physical\_description} and \texttt{visual\_notes} fields within the blueprint, combined with the overall \texttt{art\_style} directives. To further ensure consistency across various expressions, advanced prompting techniques with the T2I model can be employed to create a character sheet, which will contain the reference image of the character with multiple views.

\subsection{Stage 3: Iterative Scene Synthesis}

This stage generates each video clip corresponding to a scene in the script. It operates in an iterative loop for all scene outlines in the production blueprint.
\begin{itemize}[leftmargin=*]
    \item \textbf{Initial Frame Generation:} For each scene, an LLM determines whether to incorporate the final frame from the preceding scene. An Image-to-Image model then generates the initial frame, conditioned on the scene's textual description—encompassing setting, action, and visual parameters from the blueprint, the character references established in Stage 2 and the prior frame when applicable. This approach ensures visual coherence between successive clips while maintaining consistent character representation throughout the narrative.
    \item \textbf{Video Clip Generation:} The generated "Initial Frame" then serves as the primary input for an Image-to-Video (I2V) Model (e.g Google Veo). This I2V model, guided by the scene's action description and the established art style, generates a short video clip of about 8 seconds and 24 fps that extends the initial frame into a clip. The use of an initial frame helps maintain visual and temporal coherence in the generated clips. 
\end{itemize}

\subsection{Stage 4: Final Composition}

The final stage of the pipeline involved composing the individual video clips into a single narrative. An editing module takes the sequence of clips generated in the previous stage and concatenates them in the correct order as defined by the script. Furthermore, the \texttt{sound\_effects} and \texttt{emotional\_tone} cues guide the generation or selection of appropriate audio, which is then merged with the visual compilation. The output is the final, complete video, ready for viewing.

\section{Evaluation}

To assess the effectiveness of our proposed pipeline, we employ an evaluation strategy that covers different features of long-form T2V generation. Our goal is to measure the pipeline's ability to generate coherent videos that strictly follow the user prompts while maintaining consistency between the scenes. The evaluation combines objective automated metrics with subjective assessments from advanced AI models. Additionally, an ablation study is conducted to validate the contribution of each key component of the pipeline.

\subsection{Dataset}

\textbf{Dataset:}To evaluate our system, we collect story prompts from diverse social media sources to characterize the landscape of user-generated narrative requests. Given our focus on Indian-themed content, we prioritize prompts that incorporate Indian cultural elements, mythology, and narrative traditions. We then employ an LLM to generate variations and extensions of these prompts, expanding the candidate pool. From this augmented set, we select the 39 most representative prompts for evaluation. These prompts cover a diverse range of genres (e.g., sci-fi, fantasy, mystery, etc.), all grounded in Indian contexts and storytelling conventions. For each prompt, we generate a video using our proposed pipeline.

\subsection{Baseline Study}

To validate the architectural design of our pipeline, we compare our full model against a series of internal baselines. By systematically excluding key components, we evaluate the resulting impact on video quality and character coherence. The primary baseline configurations are:

\begin{itemize}[leftmargin=*,topsep=4pt,itemsep=4pt]
    \item \textbf{Baseline 1 (No Character Visualization):} This configuration excludes Stage 2 (Character Generation) from the pipeline. Consequently, the Image-to-Image (I2I) model generates the scene using only the textual descriptions from the blueprint and, if available, the last frame of the previous scene, without a dedicated visual reference for the character.
    
    \item \textbf{Baseline 2 (No Initial Frame):} This configuration excludes Stage 3 (I2I Seed Generation). As a result, the final Image-to-Video (I2V) model (Stage 4) receives only the textual scene descriptions from the blueprint, forcing it to operate without the visual anchor provided by the starting frame.
\end{itemize}

\subsection{Metrics}
We evaluate our system using two comprehensive metrics that assess the fidelity of the most critical components: character consistency and script adherence.

\subsubsection{Character Consistency}
We measure character consistency through a dual-pronged approach that combines both computational and semantic evaluation. Following \cite{atzmon2024multishot}, we first establish a computational baseline via cross-scene frame similarity. We uniformly sample $f$ frames from each scene, yielding $n f + 1$ frames in total (where $n$ denotes the number of scenes), combined with the reference character image from Stage 2. To isolate character features from confounding visual factors, we employ CLIPSeg \cite{luddecke2021clipseg}, a CLIP-based segmentation model, to extract the character region. We further refine this by masking the background following \cite{fu2023objectmasking}, ensuring that only the character contributes to similarity computations. We then extract deep visual features using the DINO model \cite{caron2021dino}, conditioning the feature extraction on the character's physical description specified in the blueprint. We compute pairwise cosine similarity across all frame pairs, excluding intra-frame comparisons, to accurately measure cross-scene character consistency, and average the results to obtain a quantitative consistency core. This metric is computed independently for each character.

To complement the computational metric, we additionally employ MLLM-as-a-judge evaluation, where an MLLM assesses the visual fidelity of each frame's character relative to the character's image generated in stage 2, on a scale of 0 to 10 (10 = perfect consistency). This semantic evaluation captures nuances in character preservation that may not be fully captured by feature-level similarity alone.

\subsubsection{Script and Prompt Adherence}
We assess the semantic quality of the generated videos by measuring their alignment with both the user's intent and the system's internal narrative. Following \cite{zheng2023judging} we employ a Multimodal Large Language Model (MLLM) to directly analyze the visual content of the generated video. The MLLM acts as a judge, assigning scores on a Likert scale from 1 to 5 for two distinct dimensions:

\begin{itemize}[leftmargin=*]
    \item \textbf{Script Adherence:} This metric evaluates the fidelity of the generated video to the detailed intermediate script produced by our pipeline. It ensures that specific scene details, character actions, and dialogue specified in the script are accurately visualized.
    \item \textbf{Prompt Adherence:} This metric evaluates the alignment between the final video and the original user prompt. It assesses whether the high-level intent, core concepts, and thematic elements requested by the user are preserved in the final output.
\end{itemize}

By conducting this dual-factor evaluation, we ensure that the system not only follows its own generated roadmap (the script) but also satisfies the user's original request.

\subsubsection{Bias Evaluation}
We employ specific metrics to detect potential biases in identity retention and physical stability across the generated content.

\paragraph{Subject Consistency (Identity Stability):}
To evaluate whether the model preserves the protagonist's identity across frames, we utilize the \textbf{DINOv2} architecture \cite{oquab2023dinov2}. Unlike traditional pixel-based metrics (PSNR/SSIM) which are sensitive to pose changes, DINO extracts semantic, high-level features that represent the "essence" of the subject.
Following the protocol in \cite{ruiz2023dreambooth}, we calculate the average cosine similarity between the feature embeddings of the initial anchor frame ($I_0$) and all subsequent frames ($I_t$). A score closer to 1.0 indicates robust identity retention.

\begin{equation}
    S_{subject} = \frac{1}{T-1} \sum_{t=1}^{T-1} \frac{\phi(I_t) \cdot \phi(I_0)}{||\phi(I_t)|| \cdot ||\phi(I_0)||}
\end{equation}

\paragraph{World Consistency (Physical Stability):}
To assess the stability of the background and inanimate objects—critical for immersive storytelling—we employ \textbf{CoTracker} \cite{karaev2023cotracker}, a transformer-based dense point-tracking model. We initialize a grid of tracking points across the video duration and measure their "visibility duration." Generative hallucinations (e.g., objects morphing into the background) cause tracking points to vanish. We aggregate the normalized visibility of all points to derive a World Consistency score ($S_{world}$), where a lower score indicates a "jittery" or structurally unstable environment.

\paragraph{Dynamic Degree (Motion Intensity):}
To ensure that lower stability scores are not simply artifacts of higher motion, we calculate the \textbf{Dynamic Degree} using the Farneback Optical Flow algorithm \cite{farneback2003two}. This metric quantifies the average magnitude of pixel displacement between frames.
This serves as a control variable; by stratifying our analysis based on motion intensity (Low, Medium, High), we can determine if the model is "brittle" under specific dynamic conditions, independent of the visual domain.

\section{Results and Analysis}

We present the results of our evaluation, comparing the proposed pipeline to our baselines across automated and LLM-based metrics. The findings demonstrate the significant advantages of our structured, multi-stage approach.

\subsection{Baseline Study}
\begin{table}[ht!]
\centering
\footnotesize
\setlength{\tabcolsep}{2pt}
\begin{tabularx}{\linewidth}{l
    *{4}{S[table-format=2.2]}
    S[table-format=2.2]
}
\toprule
 & {Finn} & {Felica} & {Brutus} & {Chad} & {Average} \\
\midrule
\textbf{Baseline 1 (No Stage 2)} & 6.67 & 2.83 & 6.61 & 7.00 & 5.78 \\
\textbf{Baseline 2 (No Seed)} & 0.34 & 0.08 & 1.05 & 0.67 & 0.55 \\
\textbf{Ours (Full Model)} & 7.42 & 9.00 & 6.73 & 10.00 & 7.99 \\
\bottomrule
\end{tabularx}
\caption{Character consistency comparison between our proposed method and internal baselines.}
\label{tab:baseline_results}
\end{table}

\noindent We evaluated our proposed full pipeline against two baseline configurations to isolate the contributions of specific modules. The results are summarized in Table \ref{tab:baseline_results}. All models used the identical blueprint (from Stage 1) to ensure a fair comparison.

\noindent Our analysis of the results is as follows:

\begin{itemize}[leftmargin=*]
    \item \textbf{Proposed Method (Full Pipeline):} Our complete approach achieved the highest character consistency score (7.99). This demonstrates the effectiveness of the multi-stage architecture, where a character image (Stage~2) provides a strong visual prior for the I2I seed frame, which in turn acts as a robust anchor for the final I2V animation.

    \item \textbf{Baseline 1 (No Character Image):} This baseline operates without the dedicated character image (Stage 2). Consequently, the I2I model generates both the scene and the character's appearance relying solely on textual descriptions from the blueprint. This resulted in a considerable performance gap compared to our full model (7.99~$\rightarrow$~ 5.78). While the I2I seed frame still provided a stable composition, the character's appearance was re-imagined from text for each scene, leading to inconsistencies.

    \item \textbf{Baseline 2 (No I2I Seed Frame):} This baseline excludes the I2I seed frame generation entirely. The final I2V model is forced to operate in a standard ``Text-to-Video'' mode, receiving only the textual blueprint. This led to a catastrophic drop in consistency (7.99~$\rightarrow$~ 0.55). This finding is critical: it indicates that the I2I seed frame is the most vital component for locking in character identity. Without this strong visual anchor, the I2V model struggles to maintain consistency, even when provided with a character image.
\end{itemize}

\noindent In conclusion, the comparison against these baselines validates our architectural design. The results confirm that both the dedicated character model (Stage~2) and the I2I seed frame are essential for high-fidelity character consistency.

\subsection{Character Consistency Analysis}
\begin{figure}[ht]
    \centering
    \includegraphics[width=\linewidth]{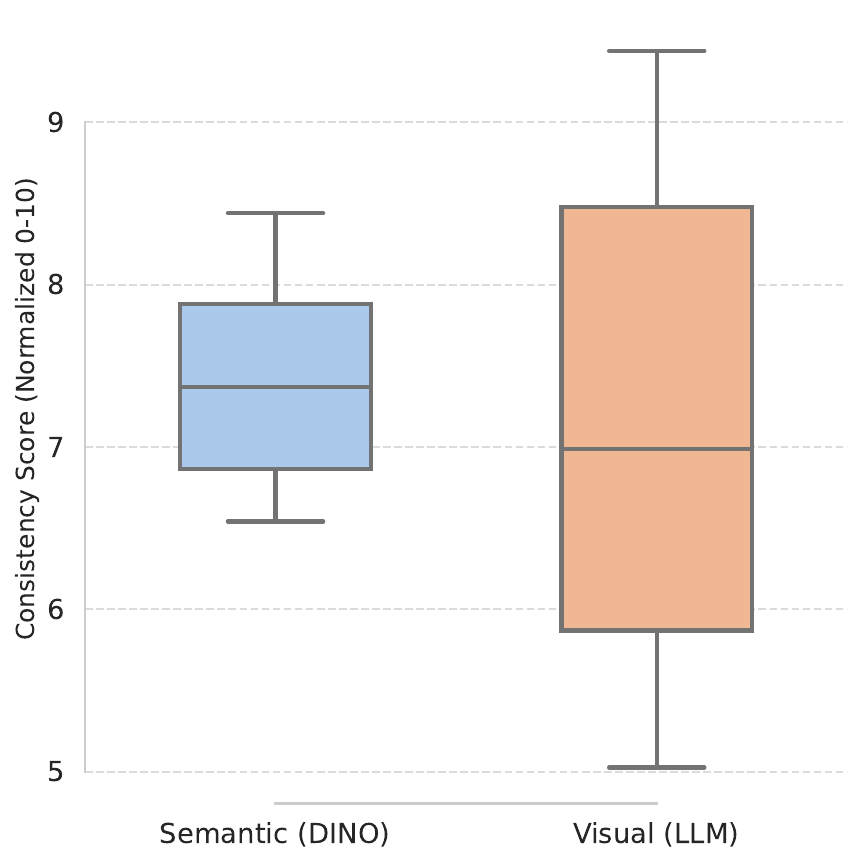}
    \caption{Distribution of Evaluation Scores.Box plots summarizing the performance across all test samples.}
    \label{fig:consistency_comparison}
\end{figure}

\paragraph{Visual Consistency:}
Our model demonstrates high visual stability,score of \textbf{7.14} ($\sigma=1.43$). The interquartile range observed in Figure \ref{fig:consistency_comparison} (red box) indicates that the model was not able to consistently maintain character features—such as clothing, facial structure, and color palette—across diverse scenes and camera angles with minimal variance.

\paragraph{Semantic Consistency:} In terms of narrative fidelity, the model achieved a mean DINO score of \textbf{7.41} ($\sigma=0.59$). The compact spread in the semantic distribution (blue box) demonstrates strong consistency in maintaining character identity across different scenes with most samples successfully maintain semantic coherence throughout the video.

Overall, the results validate that our pipeline effectively balances the trade-off between rigid visual stability and the flexibility required for semantic storytelling.

\subsection{Script and Prompt Adherence}
\begin{figure}[htbp]
    \centering
    \includegraphics[width=\columnwidth]{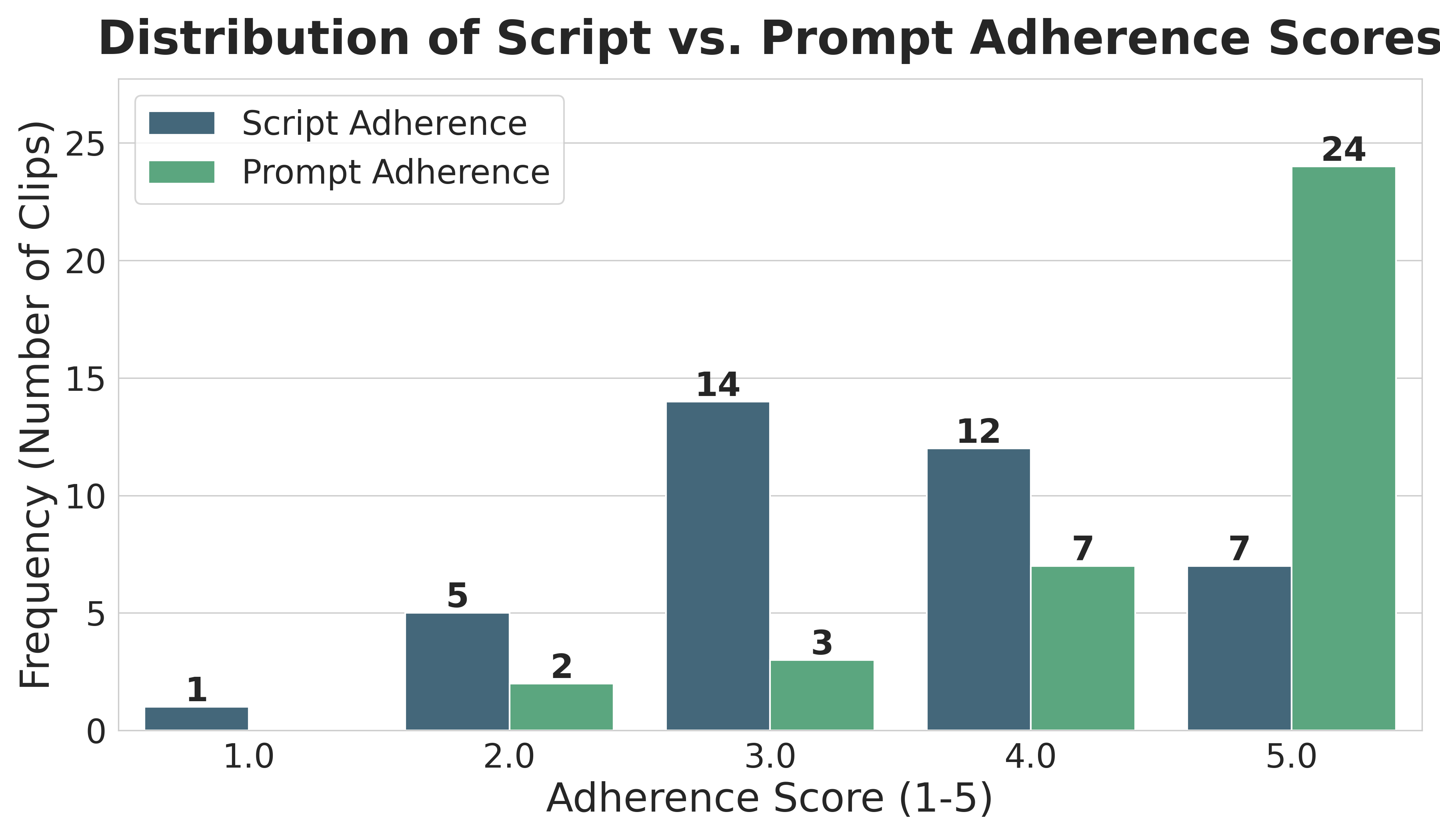}
    \caption{The distribution of scores for Script Adherence versus Prompt Adherence across all generated clips. }
    \label{fig:adherence_results}
\end{figure}
Figure \ref{fig:adherence_results} illustrates the distribution of scores for both metrics. The model demonstrated exceptional performance in Prompt Adherence, achieving a mean score of 4.47 ($SD=0.88$). A significant majority of clips (75\%) received a perfect score of 5, indicating high reliability in translating textual visual descriptions into video content.

In contrast, Script Adherence showed greater variability, with a lower mean score of 3.49 ($SD=1.02$). The distribution for script adherence is centered around scores of 3 and 4, suggesting that while the model generally follows the narrative arc, it occasionally misses specific scripted actions or timing details.

Notably, while 31 clips (approximately 86\% of the valid dataset) achieved a Prompt Adherence score of 4 or higher, only 14 clips (36\%) achieved the same threshold for Script Adherence. This disparity suggests that the current architecture prioritizes visual style transfer and prompt fidelity over complex narrative sequencing.

\subsection{Bias Analysis}
To assess the potential demographic and contextual biases in our pipeline, we conducted a comparative study between two distinct datasets: \textbf{Western/Fantasy (Control)} and \textbf{Indian/Realistic (Test)}. We evaluated performance across two dimensions: Identity Stability and Environmental Persistence.

\begin{figure}[ht]
    \centering
    \includegraphics[width=\linewidth]{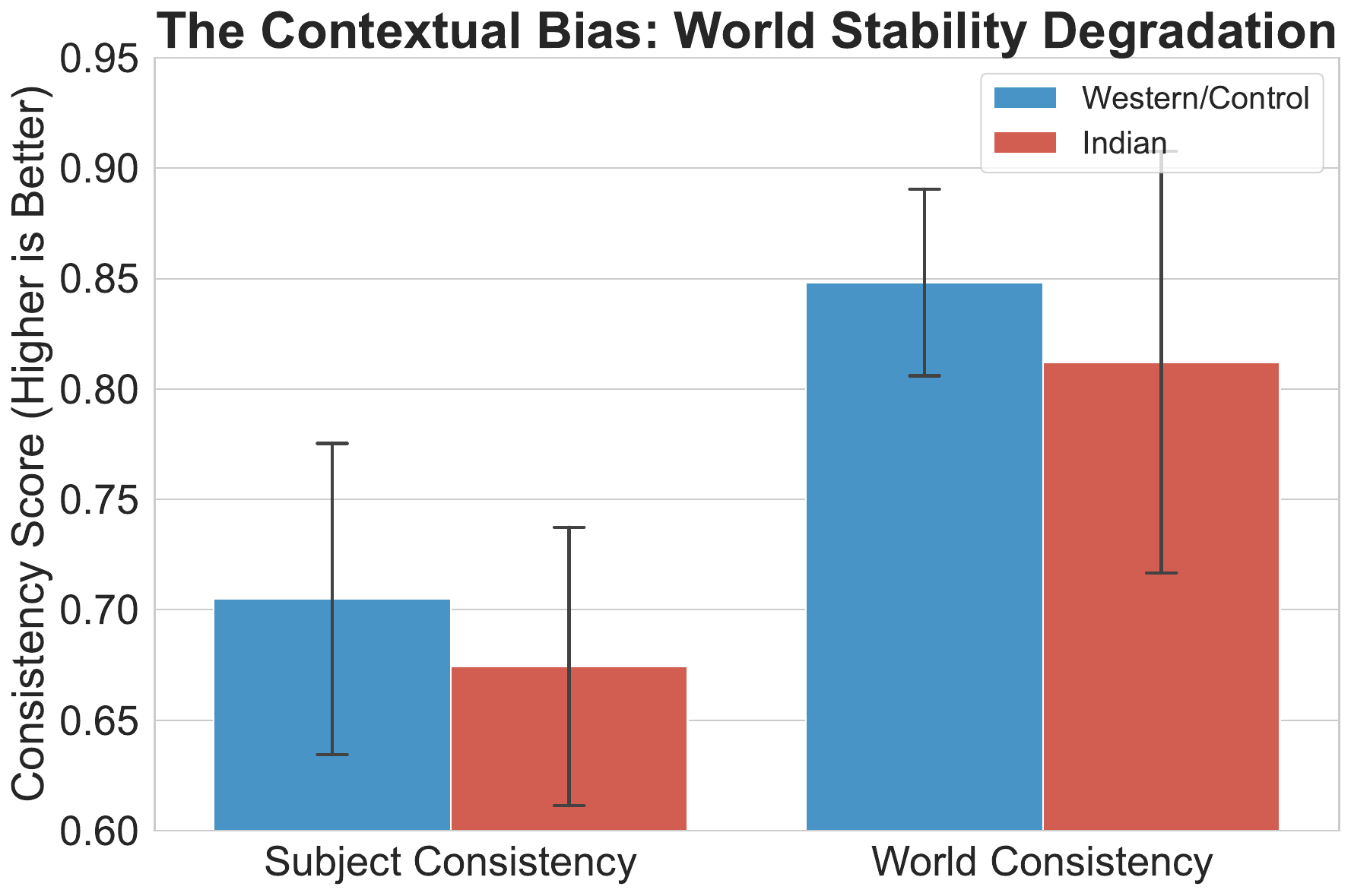}
    \caption{While Subject Consistency remains comparable across demographics, World Consistency drops significantly for Indian videos, indicating contextual fragility.}
    \label{fig:bias_gap}
\end{figure}

A critical finding of our study is the divergence between subject and world stability. As visualized in Figure \ref{fig:bias_gap}, the model exhibits a \textbf{Subject-World Decoupling} in non-Western contexts.

The \textbf{Subject Consistency gap} is considerable ($\Delta -0.03$), indicating that the generative model maintains Indian identities (faces, skin tone, clothing) with relatively less fidelity as Western identities. However, the \textbf{World Consistency gap} is significant ($\Delta -0.04$). This suggests that while the model can "anchor" a diverse character, it struggles to maintain the physical coherence of the surrounding environment (backgrounds, objects, crowds) in diverse settings.

\subsection{Stratified Motion Stress Test}
To address potential concerns regarding unequal sample sizes, we conducted a \textbf{Stratified Analysis}. We categorized all videos into three ``Difficulty Tiers'' based on their Dynamic Degree scores (Low, Medium, High) and compared performance within each tier.

\begin{figure}[ht]
    \centering
    \includegraphics[width=\linewidth]{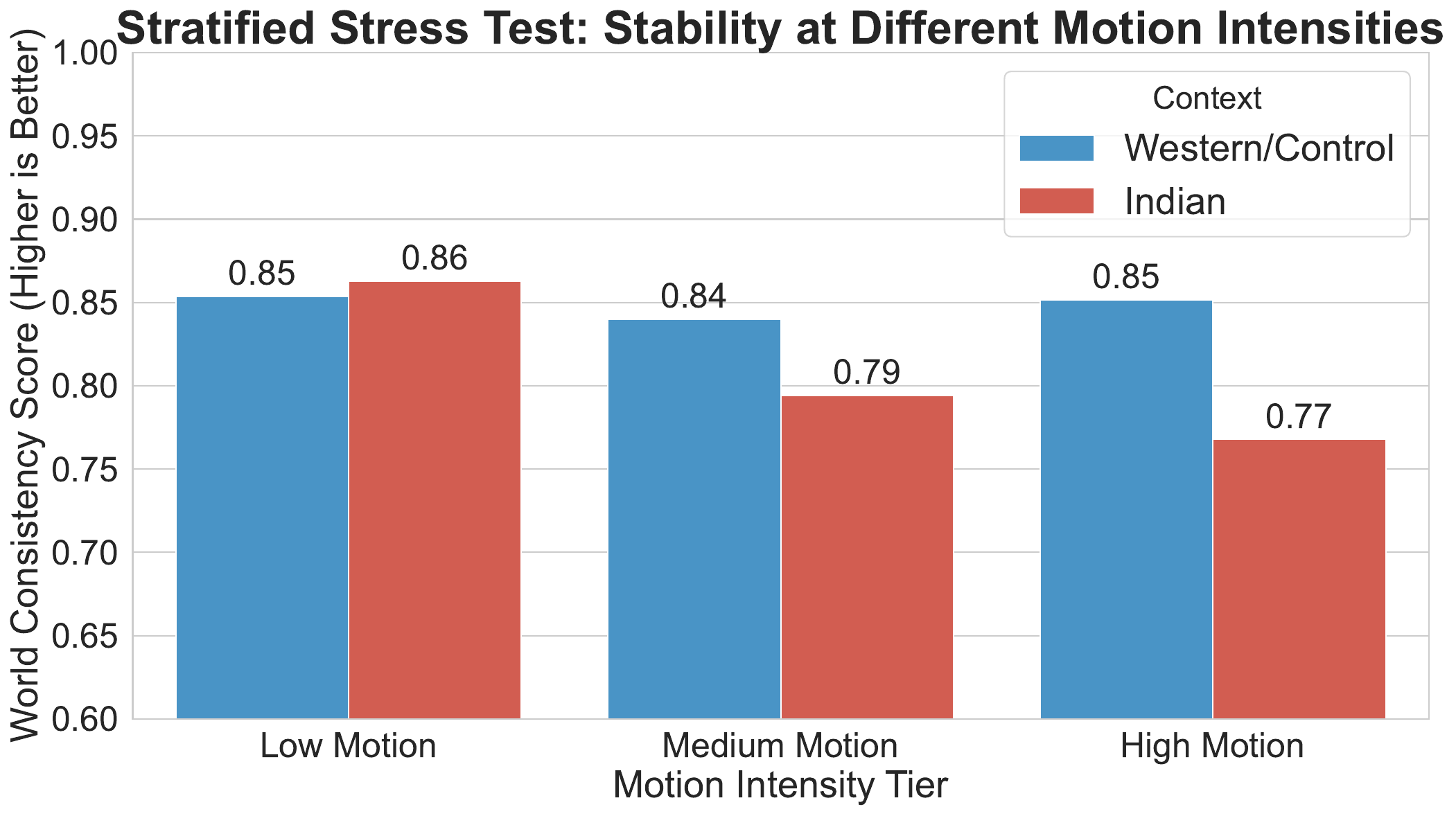}
    \caption{Stratified Analysis. The stability gap is minimal in static scenes (Low Motion) but widens drastically in dynamic scenes (High Motion), confirming that the bias is motion-dependent.}
    \label{fig:stratified_motion}
\end{figure}

The stratified results (Figure \ref{fig:stratified_motion}) confirm that the instability is not an artifact of dataset size, but a fundamental limitation of the model's physics engine in diverse contexts:
\begin{itemize}
    \item \textbf{Low Motion:} The stability gap is minimal. In static scenes, the model renders Indian environments with high fidelity.
    \item \textbf{High Motion:} The gap widens significantly. Under dynamic stress, the Western control group maintains stability, while the Indian test group degrades.
\end{itemize}

This confirms that the model possesses the \textit{capacity} to render Indian contexts but lacks the \textit{robustness} to maintain them during complex physical interactions.

\section{Conclusion}
In this work, we presented a novel "filmmaker" pipeline designed to address the challenge of creating long, cohesive video narratives. Rather than relying on end-to-end generation, our approach utilized a large language model to author detailed production scripts, followed by a character-anchoring process that guided the final scene synthesis.

Experimental results demonstrated that this structured decomposition significantly improved narrative coherence and character consistency compared to standard baselines. Our ablation study confirmed that each stage of the pipeline was essential for achieving these results. Additionally, our evaluation highlighted performance disparities in subject consistency and dynamic degree biases between Indian and Western-themed videos, underscoring the need for more culturally diverse training data in future text-to-video systems.

\section{Limitations}

While our proposed pipeline demonstrates a robust method for generating consistent and coherent short-form videos from a single prompt, we acknowledge several limitations that present opportunities for future research.

The final quality of the generated video is fundamentally capped by the performance of the underlying generative models at each stage. Failures in any stage-the script LLM, character generator, Image-to-Image model, or Image-to-Video model- will inevitably propagate and compound through the pipeline. The system's ability to render complex scenes, specific art styles or characters is therefore constrained by the training data and capabilities of these foundation models.

The blueprint specifies an action (e.g., "a character walks across the room"), but the I2V model determines the final motion path, speed, and character gait. The pipeline lacks a mechanism to explicitly direct precise cinematography (e.g., "dolly zoom," "rack focus") or complex character choreography within the video generation stage.

\section{Acknowledgement}

The authors acknowledge the use of Large Language Models (LLMs), including Gemini, ChatGPT and Claude, to assist in refining the language, improving grammatical accuracy, and enhancing the overall presentation of this paper. All outputs generated by these tools were carefully reviewed, edited, and verified by the authors. The final content remains an accurate representation of the authors' contributions.

% \begin{table}[h]
%     \centering
%     \small
%     \begin{tabular}{@{}lc@{}}
%     \toprule
%     \textbf{Evaluation Metric} & \textbf{Score} \\
%     \midrule
%     Object Presence & 4.5 / 5.0 \\
%     Action Accuracy & 3.0 / 5.0 \\
%     Setting \& Style Coherence & 2.0 / 5.0 \\
%     \midrule
%     \textbf{Overall Relevancy} & \textbf{6.0 / 10.0} Stylised \bottomrule
% \end{tabular}
% \caption{Sample scores from the LLM-as-a-Judge evaluation for a single generated video.}
% \label{tab:llm_scores}
% \end{table}

% The accompanying justification reveals the model's nuanced reas ning. For instance, the judge awarded a high score for \textit{Object Presence}, but a low score to \textit{Setting \& Style Coherence}, correctly identifying a fundamental mismatch between the prompt's request for a ``Stylized 2D/3D Hybrid'' and the video's ``detailed and realistic''  tyle. The LLM also detected subtle deviation in the narrative, such as a character slamming a control panel instead of tapping it with ''quiet resignation.'' This detailed analysis is invaluable for identifying specific failure modes and underscores the power of using LLMs for qualitative analysis.

% Bibliography entries for the entire Anthology, followed by custom entries
%\bibliography{anthology,custom}
% Custom bibliography entries onl
% --- BIBLIOGRAPHY ---
% --- BIBLIOGRAPHY ---

% This is the corrected co mand. It should not include the .bib extension.
\bibliography{custom}

\end{document}